\pdfoutput=1

\documentclass[11pt]{article}

\usepackage[]{naacl2021}

\usepackage{times}
\usepackage{latexsym}

\usepackage[T1]{fontenc} 

\usepackage[utf8]{inputenc}

\usepackage{microtype}

\usepackage{booktabs}
\usepackage{graphicx}
\usepackage{amsmath}
\usepackage{enumitem}

\title{What Will it Take to Fix Benchmarking\\ in Natural Language Understanding?}

\author{Samuel R. Bowman\\New York University\\
\texttt{bowman@nyu.edu}\And
George E. Dahl\\Google Research, Brain Team\\
\texttt{gdahl@google.com}
}

\date{}

\begin{document}
\maketitle

\begin{abstract}

Evaluation for many natural language understanding (NLU) tasks is broken: Unreliable and biased systems score so highly on standard benchmarks that there is little room for researchers who develop better systems to demonstrate their improvements. The recent trend to abandon IID benchmarks in favor of adversarially-constructed, out-of-distribution test sets ensures that current models will perform poorly, but ultimately only obscures the abilities that we want our benchmarks to measure. In this position paper, we lay out four criteria that we argue NLU benchmarks should meet. We argue most current benchmarks fail at these criteria, and that adversarial data collection does not meaningfully address the causes of these failures. Instead, restoring a healthy evaluation ecosystem will require significant progress in the design of benchmark datasets, the reliability with which they are annotated, their size, and the ways they handle social bias.
\end{abstract}

\section{Introduction}

A large and impactful thread of research on natural language understanding (NLU) has focused on improving results on benchmark datasets that feature roughly independent and identically distributed (IID) training, validation, and testing sections, drawn from data that were collected or annotated by crowdsourcing \cite{maas-etal-2011-learning,bowman-EtAl:2015:EMNLP,rajpurkar-EtAl:2016:EMNLP2016,wang2018glue}. Recent methodological progress combined with longstanding issues in crowdsourced data quality has made it so state-of-the-art systems are nearing the maximum achievable values on most of these benchmarks and thus are unlikely to be able to measure further improvements \citep{devlin-etal-2019-bert,raffel2019t5}. At the same time, these apparently high-performing systems have serious known issues and have \emph{not} achieved human-level competence at their tasks \citep{ribeiro-etal-2020-beyond}.

Progress suffers in the absence of a trustworthy metric for benchmark-driven work: Newcomers and non-specialists are discouraged from trying to contribute, and specialists are given significant freedom to cherry-pick ad-hoc evaluation settings that mask a lack of progress \citep{church_hestness_2019}.

\begin{figure}[t]
\setlength\fboxsep{0.04\columnwidth}
\centering
\noindent
\fbox{\parbox{0.91\columnwidth}{
\noindent\small
\begin{enumerate}[leftmargin=0.05\columnwidth]
\vspace{-.8em} 
    \item Good performance on the benchmark should imply robust in-domain performance on the task.\\
    $\hookrightarrow$ \textit{We need more work on dataset design and data collection methods.}
    \item Benchmark examples should be accurately and unambiguously annotated.\\
    $\hookrightarrow$ \textit{Test examples should be validated thoroughly enough to remove erroneous examples and to properly handle ambiguous ones.}
    \item Benchmarks should offer adequate statistical power.\\
    $\hookrightarrow$ \textit{Benchmark datasets need to be much harder and/or much larger.}
    \item Benchmarks should reveal plausibly harmful social biases in systems, and should not incentivize the creation of biased systems.\\
    $\hookrightarrow$ \textit{We need to better encourage the development and use of auxiliary bias evaluation metrics.}
\end{enumerate}\vspace{-.8em}
}}
    \caption{A summary of the criteria we propose.}
    \label{fig:criteria}
\end{figure}

The plight of benchmark-driven NLU research has prompted widespread concern about the assumptions underlying standard benchmarks and widespread interest in alternative models of evaluation. As an especially clear example, the documentation for the recent DynaBench benchmark suite argues that ``benchmarks saturate'', ``benchmarks have artifacts'', ``researchers overfit on benchmarks'', and ``benchmarks can be deceiving'' and use these claims to motivate abandoning the IID paradigm in favor of benchmark data that is collected adversarially by asking a broad population of annotators to try to \textit{fool} some reference neural network model.\footnote{\url{https://dynabench.org/about}} 

The DynaBench approach falls into the broader category of \textit{adversarial filtering} \citep{paperno-etal-2016-lambada,zellers2018swag,nie-etal-2020-adversarial,bras2020adversarial}. Adversarial filtering starts with a pipeline that produces candidate examples for the task, often through crowdsourcing, and then constructs a dataset by selecting those examples from the pipeline where one or more machine learning models  fails to predict the correct label. This approach is appealing in that it guarantees that, at least in the short term, existing approaches to dataset construction can be patched to keep producing data that will challenge current systems. 

However, collecting examples on which current models fail is neither necessary nor sufficient to create a useful benchmark.
Among other points of concern, this approach can create a counterproductive incentive for researchers to develop models that are  different without being better, since a model can top the leaderboard either by producing fewer errors than the adversary or by simply producing \textit{different} errors, because the examples on which these new errors would be tested will not appear in the evaluation set. One could attempt to do this by, for example, pretraining new models that deliberately avoid any data that was used to pretrain the original adversary model, in order to minimize the degree to which the idiosyncratic mistakes of the new model line up with those of the old one.  This incentive can slow progress and contribute to spurious claims of discovery.

This position paper argues that concerns about standard benchmarks that motivate methods like adversarial filtering are justified, but that they can and should be addressed directly, and that it is possible and reasonable to do so in the context of static, IID evaluation. We propose four criteria that adequate benchmarks should satisfy: benchmarks should offer a valid test of the full set of relevant language phenomena, they should be built around consistently-labeled data, they should offer adequate statistical power, and they should disincentivize the use of systems with potentially harmful biases. We then briefly survey some ongoing or promising research directions that could enable us to meet these challenges, including hybrid data collection protocols involving both crowdworkers and domain experts, larger-scale data validation, and auxiliary bias metric datasets attached to benchmarks.

\section{Background}

\paragraph{The Problem} Performance on popular benchmarks is extremely high, but experts can easily find issues with high-scoring models. The GLUE benchmark \citep{wang2018glue,nangia2019human}, a compilation of NLU evaluation tasks, has seen performance on its leaderboard approach or exceed human performance on all nine of its tasks.
The follow-up SuperGLUE benchmark project \cite{wang2019superglue} solicited dataset submissions from the NLP research community in 2019, but wound up needing to exclude the large majority of the submitted tasks from the leaderboard because the BERT model \citep{devlin-etal-2019-bert} was \textit{already} showing performance at or above that of a majority vote of human crowdworkers.
Of the eight tasks for which BERT did poorly enough to leave clear headroom for further progress, all are now effectively saturated \citep{raffel2019t5,he2020deberta}. 
State-of-the-art performance on the highly popular SQuAD 2 English reading-comprehension leaderboard \citep{rajpurkar-etal-2018-know} has long exceeded that of human annotators.

Ample evidence has emerged that the systems that have topped these leaderboards can fail dramatically on  simple test cases that are meant to test the very skills that the leaderboards focus on \citep{mccoy-etal-2019-right,ribeiro-etal-2020-beyond}.
This result makes it clear that our systems have significant room to improve. However, we have no guarantee that our benchmarks will detect these needed improvements when they're made. Most were collected by crowdsourcing with relatively limited quality control, such that we have no reason to expect that perfect performance on their metrics is achievable or that the benchmark will meaningfully distinguish between systems with superhuman metric performance. While the true upper bound on performance for any task (Bayes error) is not measurable, the fact that our systems have exceeded serious estimates of human performance leaves us with no reason to expect there to be much more headroom.

In addition, many of our best models display socially-relevant biases that render them inappropriate for deployment in many applications.\footnote{The state-of-the-art T5 model, for example, shows far more sensitivity to irrelevant gender information than humans do when making coreference judgments, according to results on the SuperGLUE leaderboard with the DNC Winogender dataset \citep{rudinger-etal-2018-gender,poliak-etal-2018-collecting}.} Our best current benchmarks do little or nothing to discourage harmful biases and, by building largely on crowdsourced or naturally-occurring text data, they likely \textit{incentivize} the development of models that reproduce problematic biases, at least to some degree.

\paragraph{The Goal} This paper lays out four criteria that we would like our benchmarks to satisfy in order to facilitate further progress toward a primarily scientific goal: \textit{building machines that can demonstrate a comprehensive and reliable understanding of everyday natural language text in the context of some specific well-posed task, language variety, and topic domain.} Among language understanding tasks, we focus on those that use labeled data and that are designed to test relatively general language understanding skills, for which the design of benchmarks can be especially difficult.

We distinguish between a \textit{task} and a \textit{benchmark}: A \textit{task}, in our terms, is a language-related skill or competency that we want a model to demonstrate in the context of a specific input--output format. A \textit{benchmark} attempts to evaluate performance on a task by grounding it to a text domain and instantiating it with a concrete dataset and evaluation metric. As a rough example, multiple-choice reading-comprehension question answering is a task, which the Cosmos benchmark \citep{cosmos} attempts to test using an accuracy metric over a specific sample of passages and questions from the English personal narrative domain. 
There is no general way to prove that a concrete benchmark faithfully measures performance on an abstract task. Nevertheless, since we can only evaluate models on concrete benchmarks, we have no choice but to strengthen the correspondence between the two as best we can.


We set aside the evaluation of computational efficiency and data efficiency, despite its relevance to many specific \textit{applications} of language technology. 
We will \textit{not} fully set aside issues of social bias. Even though it is possible for the same system to demonstrate both adept language understanding and harmful social prejudices,\footnote{The performance of models like RoBERTa \citep{liu2019roberta} or T5 \citep{raffel2019t5} on benchmarks like SuperGLUE that include some coverage of social bias is a good example of this, and typical human behavior is an even better example.} ethical concerns prompt us to argue that community-wide benchmarks should identify and disincentivize potentially harmful biases in models. The widespread sharing of trained models among NLU researchers and engineers and the fast pace of NLP R\&D work mean that it is easy for systems designed with scientific goals in mind to be deployed in settings where their biases can cause real harm. While recent initiatives around data documentation should reduce the accidental deployment of models built on inappropriate data \citep{bender-friedman-2018-data,gebru2018datasheets}, we see room to do more.

We will also set aside few-shot learning, in which tasks are made artificially difficult by training models only on small subsets of the available training data \citep[as was prominently used for GPT-3 by][]{brown2020gpt3}. This paper focuses instead on the case where one is interested in reaching excellent performance on some language task and is willing to collect data or otherwise expend resources to make that possible. While few-shot learning represents a potentially impactful direction for engineering research, and \textit{success} on some task in a few-shot setting is clear evidence of success more generally, artificial constraints on the use of training data do not fit the broad goals laid out above and do not fit many applied settings.

\section{Four Challenges}

This paper focuses on four criteria, outlined in Figure \ref{fig:criteria}, that we argue effective future benchmarks for NLU tasks should satisfy. We believe that no current benchmark for any difficult broad-domain NLU task satisfies all four:

\subsection{Validity}

If one system significantly outperforms another on some benchmark, then that result should be strong evidence that the higher-scoring system is actually better at the task tested by the benchmark. 
In other words, benchmarks are only useful for language understanding research if they evaluate language understanding. General-purpose benchmarks that are designed to cover tasks like paragraph reading comprehension over Wikipedia are only effective if they test the full range of skills that are required to understand and reason about paragraphs from Wikipedia. 

This criterion is difficult to fully formalize, and we know of no simple test that would allow one to determine if a benchmark presents a valid measure of model ability. Minimally, though, it requires the following:

\begin{itemize}
    \item An evaluation dataset should reflect  the full range of linguistic variation---including words and higher-level constructions---that is used in the relevant domain, context, and language variety.
    \item An evaluation dataset should have a plausible means by which it tests all of the language-related behaviors that we expect the model to show in the context of the task.
    \item An evaluation dataset should be sufficiently free of annotation artifacts \citep[as in][]{si2019does,Sugawara2020AssessingTB,niven-kao-2019-probing} that a system cannot reach near-human levels of performance by any means other than demonstrating the required language-related behaviors.
\end{itemize}

If a benchmark fully meets this challenge, we should expect any clear improvement on the benchmark to translate to similar improvements on \textit{any other} valid and reasonable evaluation data for the same task and language domain.\footnote{Though, of course, any model with non-zero test error could be presented with a potentially-\textit{un}reasonable benchmark entirely consisting of its own test errors.}

The rest of this section surveys common paradigms for constructing a benchmark dataset, and points to reasons that none offers a straightforward way to satisfy this criterion:

\paragraph{Naturally-Occurring Examples}

It is intuitively appealing to, where possible, build benchmark datasets based on naturally-occurring data distributions. This minimizes our effort in creating benchmarks and minimizes the risk that the benchmark is somehow skewed in a way that omits important phenomena. However, this is often not viable.

For tasks like reading comprehension or natural language inference that require multiple related texts (such as a passage and a question) as input, there is often no natural distribution that efficiently isolates the relevant task behaviors. One can find naturally-occurring distributions over questions, like those used to construct Natural Questions \citep{kwiatkowski-etal-2019-natural}, but these will generally be tied to the use contexts of a specific NLP product and will thus be limited by users' perceptions of the current abilities of that product. 

Even for single-input tasks like coreference resolution or Cloze, for which any text corpus can be the basis for a benchmark, naturalistic distributions do nothing to separate skills of interest from factual world knowledge and can be overwhelmingly dominated by the latter, making them poor metrics for incremental progress on NLU. Credible existing NLU-oriented benchmarks for such tasks  are generally heavily curated \citep{paperno-etal-2016-lambada,wsc,sakaguchi2019winogrande}.

\paragraph{Expert-Authored Examples}

Expert-constructed datasets for language understanding like FraCaS \citep{cooper1996using} and the Winograd Schema Challenge \citep{wsc} have been crucial for defining several new tasks and introducing them as objects of study. 
However, expert example construction isn't desirable for the creation of benchmarks for the use cases we focus on here.

Setting aside the logistical challenges of creating sufficiently large and diverse datasets by expert labor alone, expert authorship generally gives members of the research community direct, fine-grained control over the data on which their systems will be evaluated. Intentionally or unintentionally, this can produce data that is oriented toward linguistic phenomena that are widely studied and widely known to be important to the task at hand. While this can be helpful when building diagnostic datasets that focus on specific types of model failure \citep{cooper1996using,naik-etal-2018-stress,wang2018glue}, it is counterproductive when our goal is to build a broad-coverage benchmark dataset to set priorities and guide progress toward the solution of some task.

\citet{dunietz-etal-2020-test} and \citet{sugawarabenchmarking} work around this issue by leaning on taxonomies of required phenomena from outside NLP. This is a direction worth pursuing, but it is not clear that appropriate taxonomies will be available for most NLU tasks of interest, or that these taxonomies will be broad and thorough enough to be straightforwardly implemented as datasets.

\paragraph{Crowdsourcing}

Most recent benchmarks for language understanding have been collected, at least in part, through crowdsourcing example construction, where non-expert annotators are given some freedom to construct examples based on a simple set of guidelines. This has an obvious appeal: Using non-expert annotators significantly lowers costs and using simple guidelines significantly reduces the risk that the resulting data will be skewed artificially toward phenomena of interest to experts.

However, straightforward standard practice, as was used to collect datasets like SNLI \citep{bowman-EtAl:2015:EMNLP} and SQuAD, seem to be relatively poor at producing difficult datasets that test the intended phenomena. Existing datasets focus heavily on repetitive, easy cases and often fail to isolate key behaviors \citep{jia-liang-2017-adversarial,TSUCHIYA18.786,mccoy-etal-2019-right}.

\paragraph{Adversarial Filtering}

Given a source of examples and a model, adversarial-filtering-style approaches build a benchmark based on samples from that source for which the model fails. 
Adversarial filtering can remove examples that are easy due to trivial artifacts, but it does not ensure that the resulting dataset supports a valid test of model ability, and it can systematically eliminate coverage of linguistic phenomena or skills that are necessary for the task but already well-solved by the adversary model. This mode-seeking (as opposed to mass covering) behavior by adversarial filtering, if left unchecked, tends to reduce dataset diversity and thus make validity harder to achieve.


In contrast with this benchmark data collection setting, adversarial \textit{competitions}, in which one compares the difficulty of collecting valid task examples that are adversarial to each of several systems, could be part of a healthy evaluation ecosystem. Such an ecosystem might involve frequent formative evaluations on a conventional non-adversarial benchmark in conjunction with periodic organized evaluations in an adversarial setting.

\subsection{Reliable Annotation} 

For our benchmarks to incentivize the development of sound new methods, the labels for their test examples should be reliably correct. This means avoiding three failure cases: (i) examples that are carelessly mislabeled, (ii) examples that have no clear correct label due to unclear or underspecified task guidelines, and (iii) examples that have no clear correct label under the relevant metric due to legitimate disagreements in interpretation among annotators. The first two cases straightforwardly compromise the validity of the benchmark, but the third is somewhat subtler.

Legitimate disagreement emerges when an example can be labeled in multiple ways depending on an annotator's choice between reasonable interpretations of the text of an example. Such disagreements might stem from dialectal variants in the interpretation of words or constructions or different reasonable interpretations of the actual state of the world. As a toy example, consider the question: Does \textit{Ed ate a burrito} entail \textit{Ed ate a sandwich}? While most US English speakers would likely answer \textit{no}, many pedants and regulatory officials have argued for \textit{yes} \citep[][]{florestall2008burrito}.

When a benchmark contains many instances of this kind of legitimate disagreement, a machine learning model will be able to study a benchmark dataset's training set for clues about typical human behavior that might allow it to perform \textit{better} than any single human annotator. This effect could contribute to misleading reports of super-human performance on such benchmarks, where \textit{human performance} reflects the behavior of humans who are reporting their own judgments, rather than attempting to  predict the most frequently assigned label, as the model does.
We observe evidence of this kind of ambiguity in existing benchmarks: For example, \citet{pavlick-kwiatkowski-2019-inherent} find that 20\% of examples across several textual entailment datasets are significantly ambiguous, and \citet{kwiatkowski-etal-2019-natural} show that 36\% of short answer annotations in Natural Questions differ significantly from the majority answer.


\subsection{Statistical Power} 

Benchmark evaluation datasets should be large and discriminative enough to detect any qualitatively relevant performance difference between two models. This criterion introduces a trade-off: If we can create benchmark datasets that are both reliable and highly difficult for the systems that we want to evaluate, then moderate dataset sizes will suffice. However, if our benchmark datasets contain many examples that are easy for current or near-future systems, then we will need dramatically larger evaluation sets to reach adequate power. 

In the context of a reliable dataset that is difficult for current systems, a 1\% absolute accuracy improvement, such as that from 80\% to 81\%, may be an acceptable minimum detectable effect. In this case, an evaluation set of a few thousand examples would suffice under typical conditions seen in NLU \citep{card2020little}. Many, though not all, popular benchmark datasets satisfy this size threshold.

Since our systems continue to improve rapidly, though, we should expect to be spending more time in the long tail of our data difficulty distributions: If we build reliable datasets, much of their future value may lie in their ability to measure improvements in accuracy among highly accurate systems. For example, an improvement from 98\% accuracy to 98.1\% represents the same 5\% \textit{relative} improvement as we saw from 80\% to 81\%. To reliably detect this smaller \textit{absolute} improvement, though, requires two orders of magnitude more evaluation data \citep{card2020little}.



\subsection{Disincentives for Biased Models}

A benchmark should, in general, favor a model without socially-relevant biases over an otherwise equivalent model with such biases.
Many current benchmarks fail this test. Because benchmarks are often built around naturally-occurring or crowdsourced text, it is often the case that a system can improve its performance by adopting heuristics that reproduce potentially-harmful biases  \citep{rudinger-etal-2017-social}. Developing adequate methods to minimize this effect will be challenging, both because of deep issues with both the precise specification of what constitutes harmful bias and because of the limited set of tools that we have available to us.

There is no precise enumeration of social biases that will be broadly satisfactory across applications and cultural contexts. This can be most easily illustrated with the example of biased associations between word representations for US English \citep[as in][]{bolukbasi2016man}. Associations between race or gender and occupation are generally considered to be undesirable and potentially harmful in most contexts, and are something that benchmarks for word representations should discourage, or at least carefully avoid rewarding. If a set of word representations encodes typically Black female names like \textit{Keisha} as being less similar to professional occupation terms like \textit{lawyer} or \textit{doctor} than typically white male names like \textit{Scott} are, then a model using those representations is likely to reinforce harmful race or gender biases in any downstream content moderation systems or predictive text systems it gets used in. 

Adequately enumerating the social attributes for which we might want to evaluate bias in some context can be difficult. For example, Indian castes, like racial categories in the United States, are often signaled by names and are an axis on which managers sometimes discriminate in hiring. Caste is a salient category of social bias in India that is subject to legal and institutional recognition. However, this bias also arises in some cases within the United States, where it has no such recognition \citep{tiku_2020}, and where it could be easily overlooked by non-specialist bias researchers.

Furthermore, building such a list of attributes is also deeply political. Within living memory, popular and legal attitudes have changed significantly in the United States about attributes like race, gender, gender expression, sexual orientation, and disability. Attitudes on these issues continue to change, and new categories can gain recognition and protection over time. In many cases, this means that choosing whether to include some attribute in a computational metric of bias means choosing which group of people to align oneself with on a political issue. While there are clear ethical rules of thumb to follow when doing so,\footnote{The ACM code of ethics states, ``when the interests of multiple groups conflict, the needs of those less advantaged should be given increased attention and priority.''} making any particular choice is nonetheless likely to put researchers in conflict with established institutions in ways that can change quickly. Any strategy for handling bias in the context of NLP benchmarks will have to grapple with this difficult reality.

\section{Sketching a Solution}

Building new benchmarks that improve upon our four axes is likely to be quite difficult. Below we attempt to sketch out some possible directions for improvement along each axis.

\subsection{Improving Validity}

Building valid benchmarks will require significant new research into data collection methods, at least some of which will be specific to the task under study. We suspect that much of this work will involve improvements in crowdsourcing and the use of non-experts, as most of the annotation behind the tasks we discuss requires no expertise other than fluent knowledge of the language variety under study.

One promising direction involves methods that start from relatively high-quality crowdsourced datasets, then use expert effort to augment them in ways that mitigate annotation artifacts. The Build-it-Break-it challenge \citep{ettinger-etal-2017-towards}, the Open Reading Benchmark \citep{dua2019orb}, and the \citet{gardner-etal-2020-evaluating} contrast sets,
among their other features, allow expert annotators to add examples to a test set to fill perceived gaps in coverage or correct perceived artifacts in a starting set of crowdsourced examples. To the extent that crowdsourcing with non-experts can produce data that has broad coverage and high difficulty but retains some measurable artifacts or flaws, this compromise approach may help to create usable benchmark datasets out of the results.

Another approach brings computational linguists directly into the crowdsourcing process. This was recently demonstrated at a small scale by \citet{hu2020ocnli} with OCNLI: They show that it is possible to significantly improve data quality issues by making small interventions during the crowdsourcing process---like offering additional bonus payments for examples that avoid overused words and constructions---without significantly limiting annotators' freedom to independently construct creative examples.

Of course, implementing interventions like these in a way that offers convincing evidence of validity will be difficult. 

\subsection{Improving Handling of Annotation Errors and Disagreements}

The use of standard techniques from crowdsourcing---generally involving multiple redundant annotations for each example---can largely resolve the issue of mistaken annotations. Careful planning and pilot work before data collection can largely resolve the issue of ambiguous annotation guidelines. Handling legitimate annotator disagreements can take two fairly different approaches, depending on the goals of the benchmark.

The simplest approach treats ambiguously labeled examples in the same way as mislabeled examples, and systematically identifies and discards them during a validation phase. For some tasks, it may still be possible to test models' handling of fundamentally ambiguous linguistic phenomena or domains using \textit{unambiguous} examples: In the case of multiple-choice question answering, for example, one can construct examples where one answer candidates is only debatably correct, but all other candidates are unequivocally wrong. Any sound model would then be expected to select the debatable choice.

Alternately, one can decline to assign single, discrete labels to ambiguous examples. This can involve asking models to predict the empirical \textit{distribution} of labels that trustworthy annotators assign \citep{pavlick-kwiatkowski-2019-inherent,poesio-etal-2019-crowdsourced}, or allowing models to predict any of several answer choices that are supported by trustworthy annotators (as in the SQuAD benchmark). This comes at the cost, though, of requiring many more annotator judgments per evaluation example.

\subsection{Improving Statistical Power}

In principle, achieving adequate statistical power is straightforward: we simply estimate the number of examples required to reach the desired statistical power for any plausible short-to-medium term system evaluation for the task, and collect that number of examples. In practice, however, costs can become prohibitive.

For a relatively simple task like NLI, labeling an existing example likely requires a bare minimum of 45 seconds \citep{vania-etal-2020-asking}, and creating a new example requires at least one minute \citep{bowman2020collecting}. Even if we use these very optimistic numbers to estimate annotation speed, a ten-way-annotated dataset of 500,000 examples will still cost over \$1 million  at a \$15/hr pay rate.\footnote{This figure ignores platform fees and makes the additional optimistic assumption that only 10\% of fully-annotated examples will be discarded because of annotator disagreement.} Recruiting more experienced annotators or encouraging annotators to work more carefully could increase this figure dramatically. While such an amount of money is not completely out of reach in a well-funded field like NLP,\footnote{To put this number in context, public estimates of the cost of OpenAI's GPT-3 \citep{brown2020gpt3} exceed \$10M \citep{venturebeat2020openai}, and in machine translation, \citet{meng2019large}'s use of 512 Nvidia V100 GPUs for three months would have cost over \$1M USD on commodity cloud infrastructure.
} investments of this kind will inevitably be rare enough that they help reinforce the field's concentration of data and effort on a few high-resource languages and tasks.

For settings in which large datasets are necessary, we see no clear way to avoid high costs. Gamification, in the style of the ESP game or ZombiLingo \citep{von2004labeling,fort2014creating}, promises to offer free human labor, but at the cost of the expert time needed to refine the task definition into a game that is widely enjoyable. This approach also introduces severe constraints on the kinds of data collection protocols that can be used and raises tricky new ethical issues \citep{doi:10.1177/1056492618790921}. Ultimately, the community needs to compare the  cost of making serious investments in better benchmarks to the cost of  wasting researcher time and computational resources due to our inability to measure progress.

\subsection{Disincentives for Biased Models}

Because there is no one-size-fits-all definition of harmful social bias,
there is little prospect of creating a benchmark for language understanding that is guaranteed to never reward the development of harmfully biased models. \textbf{This is not a compelling reason to accept the status quo}, and we nonetheless have a clear opportunity to mitigate some of the potential harms caused by applied NLP systems before those systems are even developed. Opting not to test models for some plausible and potentially-harmful social bias is, intentionally or not, a political choice.

While it would be appealing to try to guarantee that our evaluation data does not itself demonstrate evidence of bias, we are aware of no robust strategy for reliably accomplishing this, and work on the closely-related problem of \textit{model} bias mitigation has been fraught with false starts and overly optimistic claims \citep{gonen-goldberg-2019-lipstick}. 

A viable alternate approach could involve the expanded use of auxiliary metrics: Rather than trying to fully mitigate bias within a single general dataset and metric for some task, benchmark creators can introduce a family of additional expert-constructed test datasets and metrics that each isolate and measure a specific type of bias. Any time a model is evaluated on the primary task test set in this setting, it would be evaluated in parallel on these additional bias test sets. This would not prevent the primary metric from unintentionally and subtly rewarding biased models, but it would combat this effect by more directly highlighting and penalizing bias in models. In addition, the fact that these metrics would target specific types of biases would make it easier for benchmark maintainers to adapt as changing norms or changing downstream applications demand coverage of additional potential harms.

For several tasks, metrics like this already exist, at least for gender in English, in the form of auxiliary test sets meant to be combined with a preexisting training set \citep{rudinger-etal-2018-gender,webster2018mind,kiritchenko-mohammad-2018-examining,li-etal-2020-unqovering}. Even so, refining these metrics and developing new ones will likely require us to face many of the same challenges that we highlight in this paper for benchmark design more generally. 

The larger challenge in implementing this approach, however, is a matter of community structure and incentive design.
Methods papers dealing with tasks for which metrics already exist rarely report numbers on these metrics. Even for the SuperGLUE benchmark, which \textit{requires} users to compute test set metrics on the DNC Winogender test set in order to reveal test set results for any other target task, a large majority of papers that report test set numbers omit this metric and decline to report potentially unflattering bias numbers \citep{raffel2019t5,pruksachatkun-etal-2020-intermediate,schick2020its,he2020deberta}.

The difficulty, then, is in developing community infrastructure to encourage the widespread reporting of metrics that address the full range of relevant likely harms. This could plausibly involve peer review norms, explicit publication venue policies, stricter versions of the SuperGLUE approach for which users can \textit{only} retrieve aggregate performance numbers, without a precise separation of the primary and bias-oriented metrics, or even the introduction of professional licensing standards.

Of course, ensuring that bias is measured and reported is not enough to prevent bias-related harms from emerging in practice: It is also necessary to ensure that those who build and deploy NLP products will take these metrics seriously and respond to them appropriately. And, of course, even if a system encodes no social bias at all, it can still be deployed in ways that produce unfair or unjust outcomes. These difficult issues are beyond the scope of a paper on benchmark design.

\section{Related Work}

The NLP and ML research communities are increasingly interested in issues surrounding data and evaluation. This section surveys relevant positions and issues that don't quite fit our schema.

\citet{welty2019metrology} advocate for the more precise reporting of the focus and abilities of test sets and metrics in ML broadly, with a focus on issues surrounding statistical power. \citet{bender-friedman-2018-data} and \citet{gebru2018datasheets} advocate for explicit freestanding datasheets documenting dataset releases of all kinds, with a focus on making potential harmful mismatches between data and application visible, and \citet{hutchinson2020accountability} argue along similar lines for a broader program of transparency and stakeholder engagement in data creation. \citet{dodge-etal-2019-show} lay out a set of best practices for results reporting, with a focus on the impact of hyperparameter tuning on model comparison. \citet{ethayarajh2020utility} advocate for the inclusion of efficiency considerations in leaderboard design. \citet{boyd-graber-borschinger-2020-question} describe ways that trivia competitions can provide a model for carefully-considered dataset design.

\citet{church_hestness_2019} revisit the arguments that motivated the NLP community's shift toward quantitative benchmarking in the early 1990s and warn that the overwhelming success of this shift has indirectly laid the groundwork for the widespread use of poor-quality benchmarks.
\citet{blodgett-etal-2020-language} challenge researchers working on social bias in NLP to focus more precisely on specific types of harm to specific populations of users, a challenge that our broad position piece does not fully meet.

NLP has had longstanding debates over the types of tasks that best test substantial language understanding skills. Many task-specific papers contribute to this debate, as does a prominent recent thread advocating for an increased focus on grounding of various kinds by \citet{bender-koller-2020-climbing}, \citet{bisk-etal-2020-experience}, \citet{zellers2020evaluating}, and others.

\section{Conclusion}

Benchmarking for NLU is broken. We lay out four major criteria that benchmarks should fulfill to offer faithful, useful, and responsible measures of language ability. We argue that departing from IID evaluation (as is seen with benchmark datasets collected by adversarial filtering) does not help to address these criteria, but lay out in broad strokes how each criterion might be addressed directly.

Nonetheless, important open research questions remain. Most centrally, it is still unclear how best to integrate expert effort into crowdsourced data collection, and we do not yet see a clear institutional model by which to ensure that bias metrics are built and used when they are most needed.

\section*{Ethical Considerations}

This paper advocates for reforms to a set of benchmarking practices that have so far largely failed to address issues of social bias, and that have thereby helped create a false sense of security among those building applied systems. While this paper offers no complete and satisfactory solutions, it proposes measures that should contribute to harm reduction.

\section*{Acknowledgments}

We thank Emily Bender, Iacer Calixto, Haokin Liu, Kyunghyun Cho, Will Huang, Jamie Kiros, Anna Rogers, and audiences at CMU, Google, and Apple for feedback on these ideas.

This project has benefited from financial support to SB by Eric and Wendy Schmidt (made by recommendation of the Schmidt Futures program), Samsung Research (under the project \textit{Improving Deep Learning using Latent Structure}), and Intuit. This material is based upon work supported by the National Science Foundation under Grant No. 1922658. Any opinions, findings, and conclusions or recommendations expressed in this material are those of the author(s) and do not necessarily reflect the views of the National Science Foundation.

\bibliography{eacl2021}
\bibliographystyle{acl_natbib}

\end{document}